\documentclass[11pt,a4paper]{article}
\usepackage[hyperref]{emnlp2020}
\usepackage{times}
\usepackage{latexsym}

\usepackage{microtype}
\usepackage{relsize}
\usepackage{soul}
\usepackage[utf8]{inputenc}
\usepackage{amssymb}
 \aclfinalcopy 
\def\aclpaperid{3210} 

\usepackage{import}

\usepackage{amsmath}
\usepackage{breqn}
\usepackage{graphicx,epstopdf}
\usepackage{multirow}
\usepackage{amsfonts}
\usepackage[english]{babel}

\usepackage[ruled,vlined]{algorithm2e}

\title{How Can Self-Attention Networks Recognize Dyck-n Languages?}

\author
       {Javid Ebrahimi, Dhruv Gelda, Wei Zhang \\
       Visa Research, Palo Alto, USA\\
       \{\texttt{jebrahim,dhgelda,wzhan}\}@visa.com\\
       }

 \usepackage{caption}
 \usepackage{cuted}

\begin{document}
\maketitle
\begin{abstract}
 
We focus on the recognition of Dyck-n ($\mathcal{D}_n$) languages with self-attention (SA) networks, which has been deemed to be a difficult task for these networks. We compare the performance of two variants of SA, one with a starting symbol (SA$^+$) and one without (SA$^-$). Our results show that SA$^+$ is able to generalize to longer sequences and deeper dependencies. For $\mathcal{D}_2$, we find that SA$^-$ completely breaks down on long sequences whereas the accuracy of SA$^+$ is 58.82$\%$. We find attention maps learned by SA$^+$ to be amenable to interpretation and compatible with a stack-based language recognizer. Surprisingly, the performance of SA networks is at par with LSTMs, which provides evidence on the ability of SA to learn hierarchies without recursion.

\end{abstract}
\section{Introduction}

There is a growing interest in using formal languages to study fundamental properties of neural architectures, which has led to the extraction of interpretable models \cite{weiss2018practical, merrill2020formal}. 
Recent work~\cite{hao2018context, suzgun2019memory,skachkova2018closing} has explored the generalized Dyck-n ($\mathcal{D}_n$) languages, a subset of context-free languages.
$\mathcal{D}_n$ consists of ``well-balanced'' strings of parentheses with $n$ different types of bracket pairs, and it is the canonical formal language to study nested structures~\cite{chomsky1959algebraic}.
 \citet{weiss2018practical} show that LSTMs \cite{hochreiter1997long} are a variant of the $k$-counter machine and can recognize $\mathcal{D}_1$ languages. The dynamic counting mechanisms, however, are not sufficient for $\mathcal{D}_{n > 1}$ as it requires emulating a pushdown automata. 
\citet{hahn2020theoretical} shows that for a sufficiently large length, Transformers \cite{vaswani2017attention} will fail to transduce the $\mathcal{D}_2$ language.
  \begin{figure} 
\centering
\includegraphics[width=1.0\linewidth]{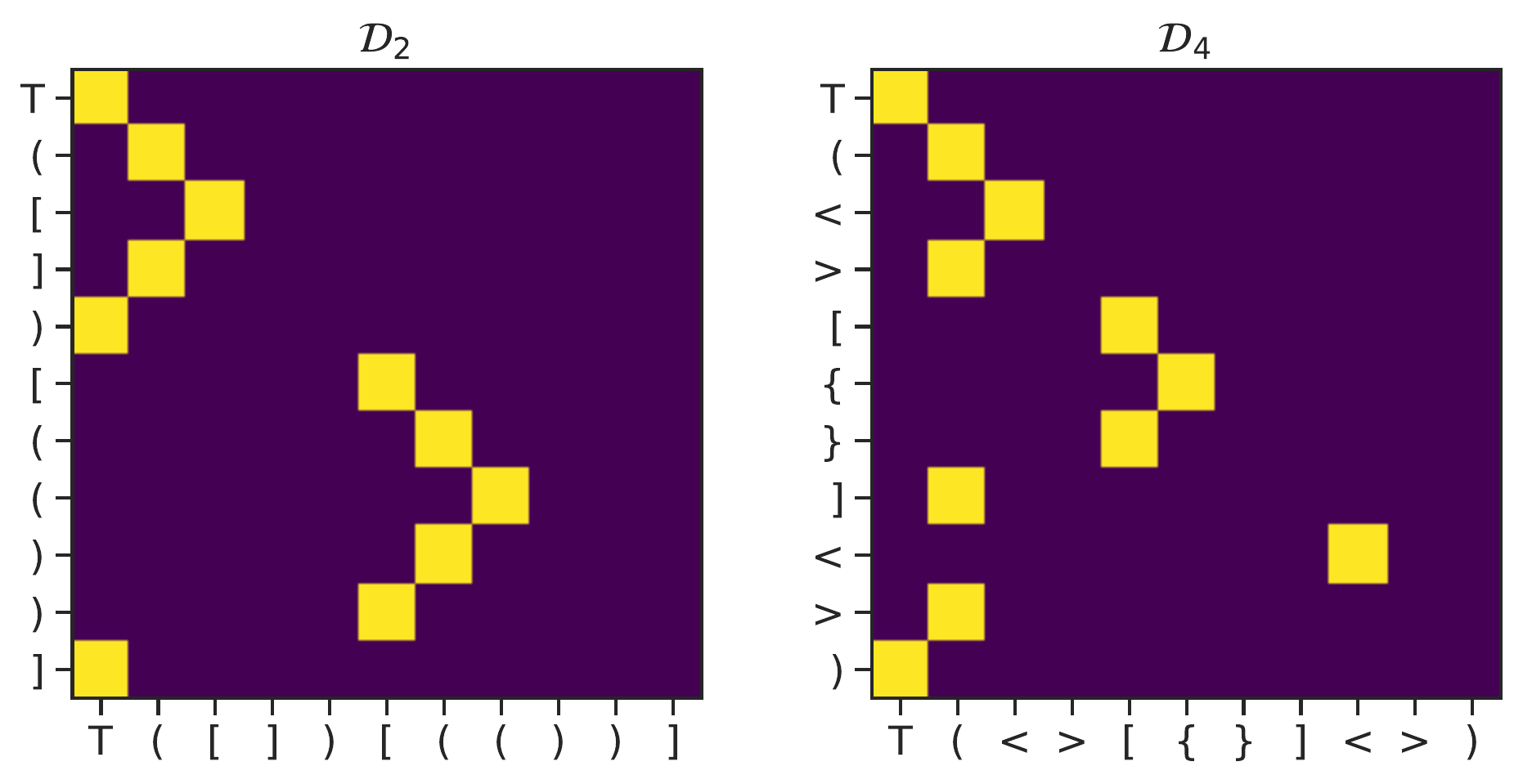}
\caption{Softmax attention scores of the second layer of a suffix-masked SA$^+$, for a $\mathcal{D}_2$ and a $\mathcal{D}_4$ sequence. The rows and columns denote queries and keys, respectively. The layer produces virtually hard attentions, in which each symbol attends only to one preceding symbol or itself. The attended symbol is either the starting symbol (T) or the last unmatched opening bracket.}
\label{fig:dyck3}
\end{figure}

We empirically show that with the addition of a starting symbol to the vocabulary, 
a two-layer multi-headed SA network (i.e., the encoder of a Transformer) is able to learn $\mathcal{D}_n$ languages, and generalize to longer sequences, although not perfectly. As shown in Figure \ref{fig:dyck3}, the network is able to identify the corresponding closing bracket for an opening bracket, in what resembles a stack-based automaton. For example, the symbol ``]'' in the string ``([])'', will first pop ``['' from the stack, then it attends to ``('', the last unmatched symbol, which will determine the next valid closing bracket.  The starting symbol (T) enables the model to learn the occurrence of the end of a clause
or the end of the sequence, which can be regarded as a mechanism to represent an empty stack.
 
Our work is the first to perform an empirical exploration of SA on formal languages. We present detailed comparison between an SA which incorporates a starting symbol (SA$^+$), and one that does not (SA$^-$), and demonstrate significant differences in their generalization across the length of sequences and the depth of dependencies. 

Recent work has suggested that the ability of self-attention mechanisms to model hierarchical structures is limited.
\citet{shen2019ordered} show that the performance of Transformers on tasks such as logical inference~\cite{bowman2015large} and ListOps~\cite{nangia2018listops} is either poor or worse than LSTMs. \citet{tran2018importance} have also reported similar results on SA, concluding that recurrence is necessary to model hierarchical structures. In comparison, our results show that SA$^+$
outperforms LSTM on $\mathcal{D}_n$ languages except for $\mathcal{D}_2$ on longer sequences. 
\citet{papadimitriou2020pretraining} posit that the ability of neural models to learn hierarchical structures can be attributed to a ``looking back'' capability, rather than directly encoding hierarchies. Our analysis sheds light on the ability of SA to learn hierarchical structures by \textit{elegantly} attending to the correct preceding symbol. 
\section{Related Work}
Formal languages such as $a^nb^n, a^nb^nc^md^m$ (context-free) and $a^nb^nc^n, a^{n+m}b^nc^m$ (context-sensitive) have been extensively studied and recognized using RNNs~\cite{elman1990finding, das1992learning, steijvers1996recurrent}. 
But the performance of same recurrent architectures on $\mathcal{D}_n$ languages is poor and suffers from the lack of generalization. 
\citet{sennhauser2018evaluating} and \citet{bernardy2018can} study the capability of RNNs to predict the next possible closing parenthesis at each position in the $\mathcal{D}_n$ string and found that the generalization at higher recursion depths is poor. \citet{hao2018context} reported that stack-augmented LSTMs achieve better generalization on $\mathcal{D}_n$ languages but the network computation does not emulate a stack. More recently,~\citet{suzgun2019memory} proposed memory-augmented recurrent neural networks and defined a sequence classification task for the recognition of $\mathcal{D}_n$ languages. 
\citet{yu2019learning} explored the use of attention-based seq2seq framework for $\mathcal{D}_2$ languages and found that the generalization to sequences with higher depths is still lacking. 
Besides empirical investigations, formal languages have been studied theoretically for understanding the complexity of neural networks~\cite{siegelmann1992computational,perez2019turing}, mostly under assumptions that cannot be met in an experiment-- infinite precision or unbounded computation time. 
\section{Experiments}
We follow prior works \cite{gers2001lstm,suzgun2019memory}, and formulate the recognition of $\mathcal{D}_n$ languages as a transduction task: Given a valid string, we ask the model to predict the next possible symbols auto-regressively. To illustrate, consider an input string ``[~(~)~]~(~['' in the $\mathcal{D}_2$ language, we seek to predict the set of next valid brackets in the string-- (, [, or ]. We consider an input to be accurately recognized only if the model correctly predicts the set of all possible brackets at each position in the input sequence. Throughout the paper, we refer to a \textit{clause} as a substring, in which the number of closing and opening brackets of each type of bracket are equal. 

We train two multi-headed self-attention networks (i.e., only the encoder part of a Transformer), one of which incorporates an additional starting symbol in the vocabulary (SA$^+$), and the other does not (SA$^-$). 
For each model, the number of layers is 2, the number of attention heads $h$ = 4 and model dimension $d$ = 256. We use learnable embeddings
to convert each input symbol to a 256-dimensional vector. We also add residual connections around each layer followed by layer normalization, similar to the standard Transformer ~\cite{vaswani2017attention}. We train two unidirectional LSTMs, one with the starting symbol (LSTM$^+$) and the other without it (LSTM$^-$). The LSTMs use 320-dimensional hidden states and a 320-dimensional vector for learned input embeddings. Our SA and LSTM variants all have around 1.6M parameters\footnote{We found dropout to be detrimental to the performance, and hence we removed it from all models.}. 
We use Adam~\cite{kingma2014adam} for optimization. For SA$^+$ and SA$^-$, we vary the learning rate $\eta$ as 
\begin{equation}
\eta = \text{const}\cdot\text{min}(\text{itr}^{-0.5}, \text{itr} \cdot \text{warmup}^{-1.5}),
\end{equation}
where itr refers to the iteration number and warmup is set to 10k. We tuned the hyper-parameter const, using the values [0.01, 0.1, 1.0, 10], and used 0.1. For LSTMs, we use an initial learning rate of 0.001 but with no learning rate scheduling.

\iftrue
\begin{table*}[h]
    \centering
\scalebox{0.8}{
\begin{tabular}{|c|c|c||c|c||c|c||c|c|}
\hline
\multirow{2}{*}{Model} & \multicolumn{2}{c|}{$\mathcal{D}_1$} & \multicolumn{2}{c|}{$\mathcal{D}_2$}  & \multicolumn{2}{c|}{$\mathcal{D}_3$} & \multicolumn{2}{c|}{$\mathcal{D}_4$}  \\
\cline{2-9}
 & 76-100  & 102-126 &     76-100  & 102-126  &  76-100  & 102-126  &  76-100  & 102-126      \\
 \hline
 SA$^{-}$ & 100.0  & {98.88} & 14.52 & 0.006 & 32.62 & 5.50 &  42.94  &  9.080 \\
 \hline
 SA$^{+}$ & 100.0  & 100.0 & \textbf{93.34}  & {58.82}  & \textbf{93.18} &  \textbf{66.88} & \textbf{93.78}   & \textbf{72.38} \\
 \hline
  LSTM$^{-}$ & 100.0  &  99.64  & 88.30  & \textbf{73.20}  & 85.16 &  65.06 & 78.92   &  60.24 \\
 \hline
  LSTM$^{+}$ & 100.0  &  100.0  & 87.00  & {70.90}  & 82.44 &  63.56  & 76.66   &  55.90 \\
 \hline
 
\end{tabular}
}
    \caption{Performance of SA and LSTM variants on Dyck-n languages for different sequence lengths.}
    \label{tab:accuracy}
\end{table*}
\fi
\begin{figure}
    \centering
    \scalebox{1.0}{
    \includegraphics[width=\linewidth]{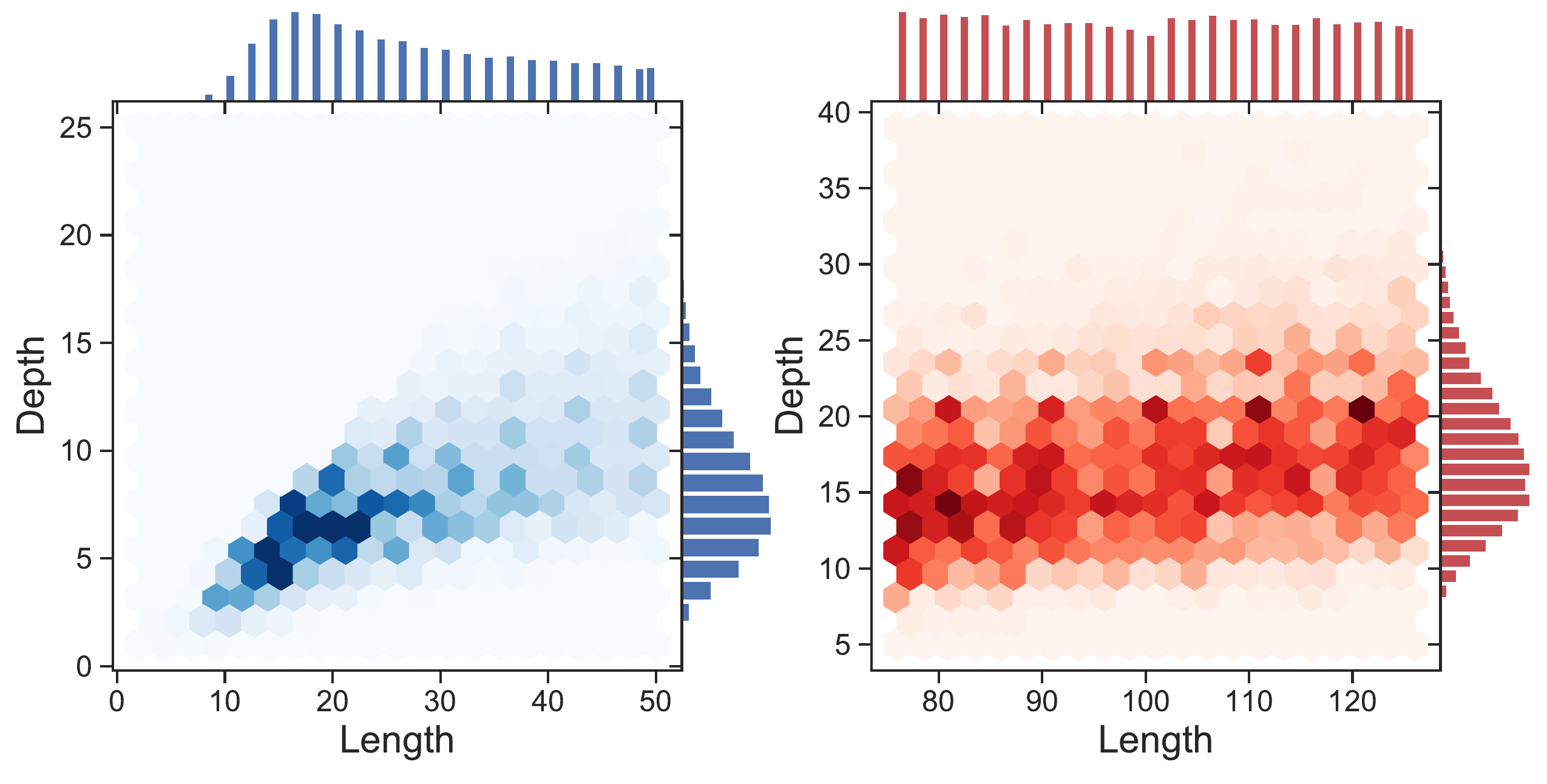}
    }
    \caption{Joint distribution of $\mathcal{D}_2$ language based on the length and depth of sequences in training (blue) and evaluation (red). The top and right axes also show the marginal distribution for length and depth respectively. 
    }
    \label{fig:hist}
\end{figure}
 
We re-generate the synthetic dataset for our experiments through the probabilistic context-free grammar (PCFG) already described in the existing literature~\cite{suzgun2019memory}. For instance, the PCFG for Dyck-2 language can be defined as: (1) $S \xrightarrow[]{}[S]$, (2) $S \xrightarrow[]{}\{S\}$, (3) $S \xrightarrow[]{}SS$, and (4) $S \xrightarrow[]{}\varepsilon$, each with probability $p= 0.25$. For each $\mathcal{D}_n$ language, we train on 32k sequences of length 2-50, validate on 3.2k sequences of length 52-74, and evaluate on 10k sequences divided equally over the length intervals 76-100 and 102-126.

Figure \ref{fig:hist} shows the distribution of length and depth of $\mathcal{D}_2$ sequences in training and evaluation. 
For higher Dyck languages ($\mathcal{D}_{n>2}$), the training and evaluation datasets have similar depth and length distributions because the PCFG give equal probability to different pairs of parentheses and the total probability for rules of the form $S\xrightarrow[]{}(S)$, $S\xrightarrow[]{}[S]$, ... is 0.5. We perform experiments on $\mathcal{D}_{1}$, $\mathcal{D}_2$, $\mathcal{D}_3$, and $\mathcal{D}_4$ languages. Note that the number of pairs of parentheses cannot be increased arbitrarily without requiring modifications to the experimental setup: We varied the length of sequences during training from 2 to 50, which could contain at most 25 different pairs.

In our sequence prediction task, the input vocabulary ($V_n^i$) for a $\mathcal{D}_n$ language consists of 2$n$+1 symbols: $n$ pairs of brackets (or parentheses), and an additional starting symbol T whereas the output vocabulary ($V_n^o$) does not include the starting symbol T.  
 Since there might exist multiple possibilities for the next bracket in a sequence, we adopt a multi-label classification approach wherein the outputs are encoded as a $k$-hot vector and the network is optimized using the binary cross-entropy loss function given by
\begin{equation}
\scalebox{1.25}{$\mathcal{L}$} = \mathlarger{\sum\limits_{i=1}^{|V^o_n|}} \Big \{ \hat{y_{i}}~\log(y_{i}) + (1-\hat{y_{i}})~\log(1-y_{i}) \Big \}  ,
\end{equation}
where $|V^o_n|$ is the output vocabulary size (2 for $\mathcal{D}_1$, 4 for $\mathcal{D}_2$, 6 for $\mathcal{D}_3$, 8 for $\mathcal{D}_4$), $\hat{y_{i}} \in \{0,1\}$ and $y_i$ are the target and prediction for label $i$, respectively.

\subsection{Evaluation}

\begin{figure*}[h]
    \centering
    \scalebox{0.85}{
    \includegraphics[width=\textwidth]{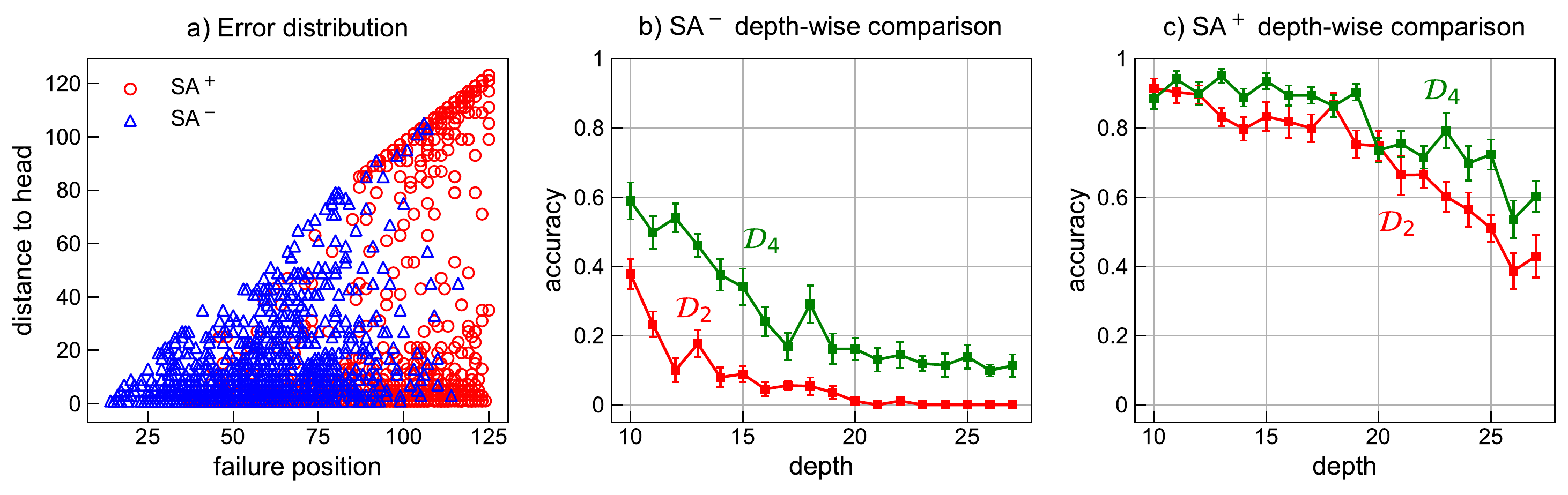}
    }
    \caption{In a, we plot the distribution of the errors made by SA$^+$ and SA$^-$, based on the position of the mispredicted symbol, and its distance to its head. In b and c, we plot the performance of the models as depth increases.}
    \label{fig:fail}
\end{figure*}
 
Table \ref{tab:accuracy} compares the accuracy of SA$^+$ and SA$^-$ on $\mathcal{D}_{1}$, $\mathcal{D}_2$, $\mathcal{D}_3$, and $\mathcal{D}_4$ languages. For both models, the performance on $\mathcal{D}_1$ is almost perfect ($>98\%$) and does not show any degradation with increase in sequence length. The accuracy of SA$^-$ on $\mathcal{D}_2$ is 14.52$\%$ for sequences with length 76-100 and completely fails beyond it. In comparison, the performance of SA$^+$ on $\mathcal{D}_2$ is significantly better, 93.34$\%$ and 58.82$\%$ for sequences of length 76-100 and 102-126, respectively. 
The performance of SA$^-$ improves on $\mathcal{D}_3$ and $\mathcal{D}_4$, compared to $\mathcal{D}_2$, with an accuracy of 32.62$\%$ and 42.94$\%$, respectively for sequences of length 76-100. 
The performance of SA$^+$ is nearly constant ($\sim$93$\%$) on $\mathcal{D}_{n\geq2}$ for sequences of length 76-10 but there is significant improvement from $\mathcal{D}_2$ (58.82$\%$) to $\mathcal{D}_3$ (66.88$\%$) and $\mathcal{D}_4$ (72.38$\%$) for sequences of length 102-126.

Unlike SA, the performance of LSTM degrades after the addition of the starting symbol, with the biggest drop (4.3\%) on $\mathcal{D}_4$ for sequence length of 102-106. The starting symbol has enabled SA to attend to the correct preceding token, but it has been ineffective for LSTM. For $\mathcal{D}_2$ sequences of length 102-126, LSTM$^{-}$ achieves an accuracy of 73.20$\%$, an improvement of $\sim$14$\%$ over SA$^+$. On all other comparisons, SA$^{+}$ outperforms LSTM$^-$. 

We observe another interesting distinction between the two architectures. The accuracy of LSTM deteriorates as the number of pairs of brackets increases, while the accuracy of SA$^+$ and SA$^-$ improves. To understand this phenomenon, we looked at the training, validation, and test sets of each language, and found that while validation and test sets of each $\mathcal{D}_n$ language almost always ($>99\%$) includes sequences of $n$ different brackets, the training set could include sequences of $1\leq m<n$ types of brackets. This implies that SA benefits from data augmentation with sequences from other languages, and LSTM does not. Put differently, these results suggest LSTM has a strong inductive bias, perhaps in counting \cite{kharitonov2020they}, which might result in degradation of its performance in higher Dyck languages.

\begin{algorithm}
\footnotesize
\texttt{\textbf{import} numpy \textbf{as} np}\\
\vspace{0.2cm}
\texttt{\textbf{def} get_match(seq, opening=\textquotesingle([\textquotesingle):}\\
 \hspace{.2cm}\texttt{stack, match = [], len(seq)*[-1]}\\ 
 \hspace{.2cm}\texttt{\textbf{for} idx, s \textbf{in} enumerate(seq):}\\
    \hspace{.4cm}\texttt{\textbf{if} s \textbf{in} opening:}\\
        \hspace{.6cm}\texttt{stack.insert(0, idx)}\\
    \hspace{.4cm}\texttt{\textbf{elif} s != \textquotesingle T\textquotesingle:}\\
        \hspace{.6cm}\texttt{stack.pop(0)}\\
        \hspace{0.6cm}\texttt{\textbf{if} len(stack) > 0:}\\
            \hspace{0.8cm}\texttt{match[idx] = stack[0]}\\
 \hspace{.2cm}\texttt{\textbf{return} match}\\
 \vspace{0.1cm}
\texttt{\textbf{def} is_compatible(seq, atten_map}):\\
 \hspace{.2cm}\texttt{match = get_match(seq)}\\
 \hspace{.2cm}\texttt{\textbf{for} idx, m \textbf{in} enumerate(match):}\\
    \hspace{0.4cm}\texttt{p = np.argmax(atten_map[idx])} \\   
    \hspace{.4cm}\texttt{\textbf{if} m != p \textbf{and} m != -1:}\\
        \hspace{.6cm}\texttt{\textbf{return False}}\\
 \hspace{.2cm}\texttt{\textbf{return}} \texttt{\textbf{True}}\\
  \caption{Compatibility of an attention map with a stack-based recognizer.}
 \label{alg:match}
\end{algorithm}

\normalsize 
\begin{figure*}[t]
    \centering
    \scalebox{0.9}{
        \includegraphics[width=\textwidth]{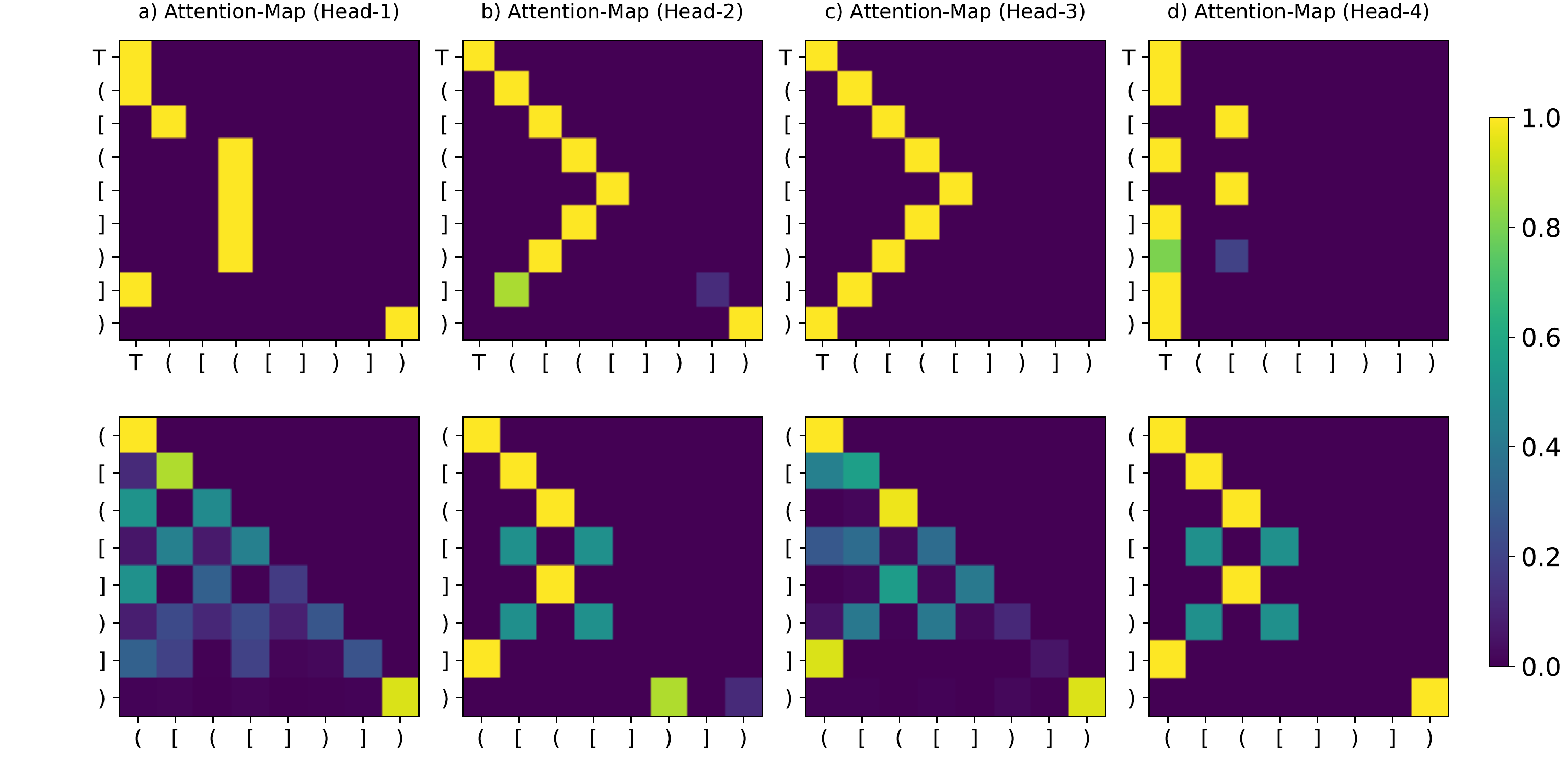}
        }
    \caption{Comparing SA$^+$ (top) and SA (bottom), based on their attention maps on a $\mathcal{D}_2$ sequence. The third head of SA$^+$ has produced weights that are compatible with the operations of a stack-based recognizer.}
    \label{fig:vis}
\end{figure*}

\subsection{Error Analysis}

We define failure position ($f_p$) as the position of the first symbol in the sequence where the model failed to correctly predict the next set of possible parentheses,
For each symbol in a $\mathcal{D}_n$ sequence: (i) depth ($d_p$) is the number of unmatched parenthesis up to and including that symbol, and (ii) distance to head ($d_h$) is the number of symbols between the mis-classified closing bracket and its opening counterpart. 
Figure~\ref{fig:fail}a plots the error distribution of SA$^+$ and SA$^-$ in terms of failure position ($f_p$) and distance to head ($d_h$). There is a clear separation between the two models in terms of what ``types'' of errors are made. SA$^-$ breaks quite early on in the sequence, with majority of the errors occurring at $f_p$ = 25-75 whereas whereas the errors of SA$^+$ are mostly concentrated at $f_p >$ 80. Figure~\ref{fig:fail}b-c shows how the performance of SA$^+$ and SA$^-$ change with depth ($d_p$) for $\mathcal{D}_2$ and $\mathcal{D}_4$ languages. SA$^-$ is very sensitive to depth as the accuracy decreases rapidly for $\mathcal{D}_2$ from $\sim$38$\%$ at $d_p=10$ to a complete failure beyond $d_p=20$. In comparison, the drop in accuracy for SA$^+$ is less severe, $\sim$ 94$\%$ at $d_p=10$ to $\sim$ 72$\%$ at $d_p=20$.

\begin{figure}
    \centering
    \includegraphics[width=0.75\linewidth]{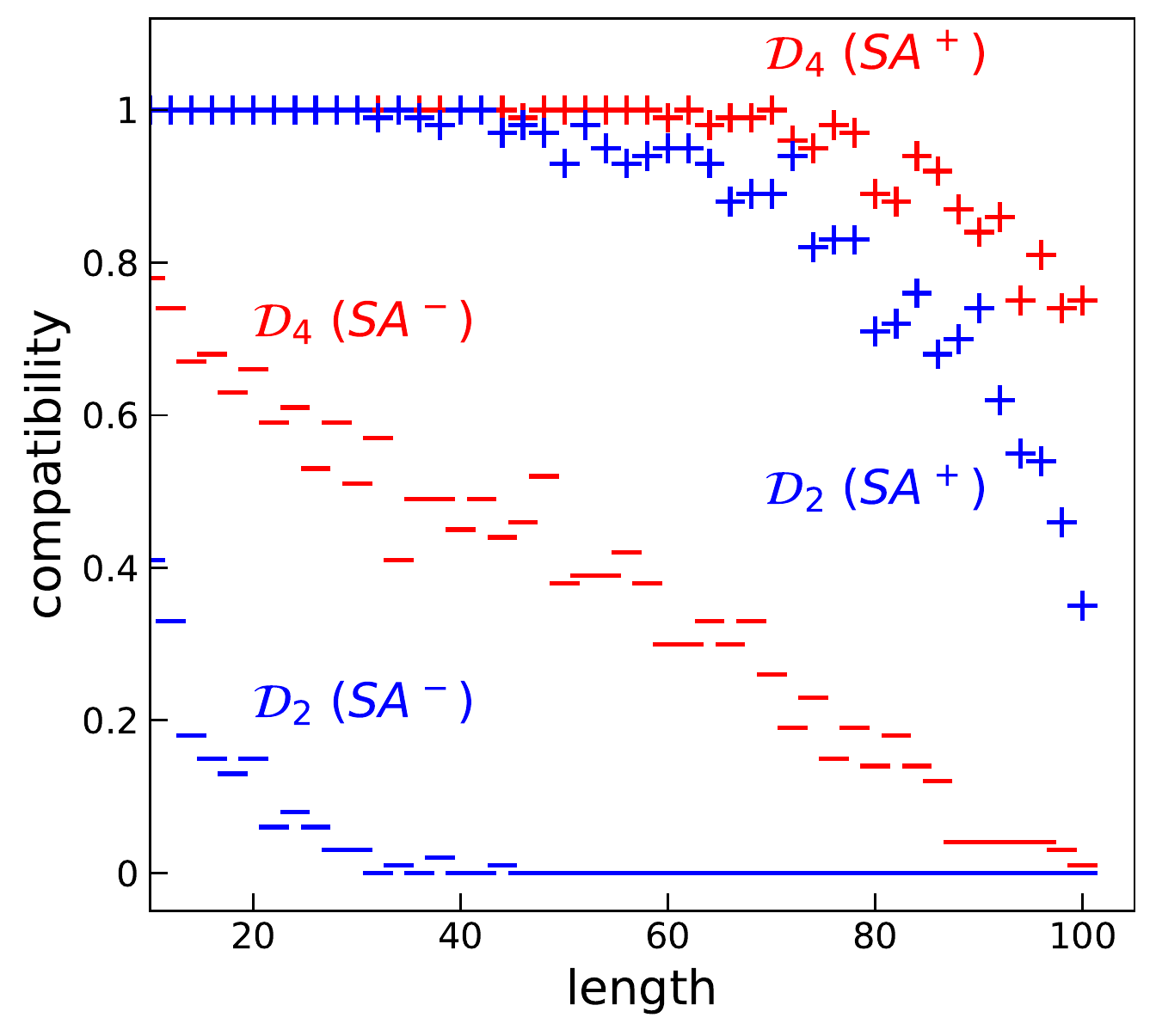}
    \caption{Compatibility versus length for SA$^+$ and SA$^-$ on $\mathcal{D}_2$ and $\mathcal{D}_4$ languages.}
    \label{fig:interpret}
\end{figure}

\section{Compatibility With a Stack-Based Recognizer}
The ability of (memory-less) SA networks to recognize $\mathcal{D}_{n>1}$ languages is intriguing. 
In this section, we contrast second-layer attention maps produced by SA$^+$ and SA$^-$, and provide insights into the underlying mechanism which leads to the success of $\text{SA}^+$.

We define \textit{compatibility} as a quantitative measure for the alignment of the state of a stack-based language recognizer ($M$) with the attention maps. $M$ has access to the top of a hypothetical stack, and can push and pop depending on the opening and closing brackets, respectively. 
Based on this analogy, all opening brackets should attend to themselves, and all closing brackets should first do a pop, and then attend to the last unmatched bracket. For example, the symbol ``]'' in the string ``([])'', will first pop ``['' from the stack, then it attends to ``('', the last unmatched symbol, which will determine the next valid closing bracket. If for every closing symbol in the sequence, the highest attention score of at least one of the heads points to the correct bracket, then we consider the SA compatible. Furthermore, for a fair comparison between SA$^+$ and SA$^-$, we do not push the starting symbol to the stack and only consider closing brackets which are not at the end of a clause.

Figure \ref{fig:interpret} plots the compatibility of SA$^+$ and SA$^-$ versus sequence length. We find that SA$^{-}$ on $\mathcal{D}_2$ has almost zero compatibility, even for sequence lengths seen during training (40-50), on which it achieves close-to-perfect accuracy. In comparison, SA$^{+}$ has perfect compatibility for sequence lengths seen during training, and maintains a high degree of compatibility for longer ones. Further, perhaps not surprisingly, the Pearson correlation between the distribution of accuracy and compatibility across lengths 50-100 is $\gtrsim$ 90\% for all SA$^+$ models.  

Figure~\ref{fig:vis} shows the attention maps of all four heads of $\text{SA}^+$ and $\text{SA}^-$ for the $\mathcal{D}_2$ sequence ``([([])])''. We observe that the third head of SA$^+$ matches our expectation of a stack-based recognizer.
An important feature of the third head is that the last symbol attends to the starting symbol T.  The starting symbol has enabled the model to learn the occurrence of the end of a clause and the end of the whole sequence.

\section{Conclusion and Future Work}
 
We provide empirical evidence on the ability of self-attention (SA) networks to learn generalized $\mathcal{D}_n$ languages.
We compare the performance of two SA networks, SA$^+$ and SA$^-$, which differ only in the inclusion of a starting symbol in their vocabulary. 
We demonstrate that a simple addition of the starting symbol helps SA$^+$ generalize to sequences that are longer and have higher depths. 
The competitive performance of SA (no-recurrence) against LSTMs might seem surprising, considering that the recognition of $\mathcal{D}_n$ languages is an inherently hierarchical task. From our experiments, we conclude that recognizing Dyck languages is not tied to recursion, but rather learning the right representations to look up the head token.
Further, we find that the representations learned by SA$^+$ are highly interpretable and the network performs computations similar to a stack automaton. Our results suggest formal languages could be an interesting avenue to explore the interplay between performance and interpretability for SA. Comparisons between SA and LSTM reveal interesting contrast between the two architectures which calls for further investigation. Recent work \cite{katharopoulos2020transformers} shows how to express the Transformer as an RNN through linearization of the attention mechanism, which could lay grounds for more theoretical analysis of these neural architectures (e.g., inductive biases and complexity.)

\bibliography{emnlp2020}
\bibliographystyle{acl_natbib}

\newpage
\noindent
\setcounter{section}{0}
%
%

\def\thesection{\Alph{section}}



\noindent
{\huge \textbf{Appendix}}

\section{More on the Starting Symbol} 
We have observed a significant difference in the generalization and interpretability of SA$^+$ and SA$^-$ models. But it is not obvious {how} the addition of a starting symbol can cause such a change. We first point out that recognizing $\mathcal{D}_1$ is trivial, and a one-layer SA (with or without the starting symbol) achieves perfect accuracy on this task. In addition, a two-layer SA$^+$ is not interpretable for $\mathcal{D}_1$, as the task does not require learning to attend to the correct preceding token, but rather a simple counting mechanism suffices. Further, we found that two layers of SA are necessary for the recognition of $\mathcal{D}_{n>1}$ and the addition of more layers does not improve generalization, but rather degrades it. 

 \begin{strip}
\centering
  \includegraphics[width=\textwidth]{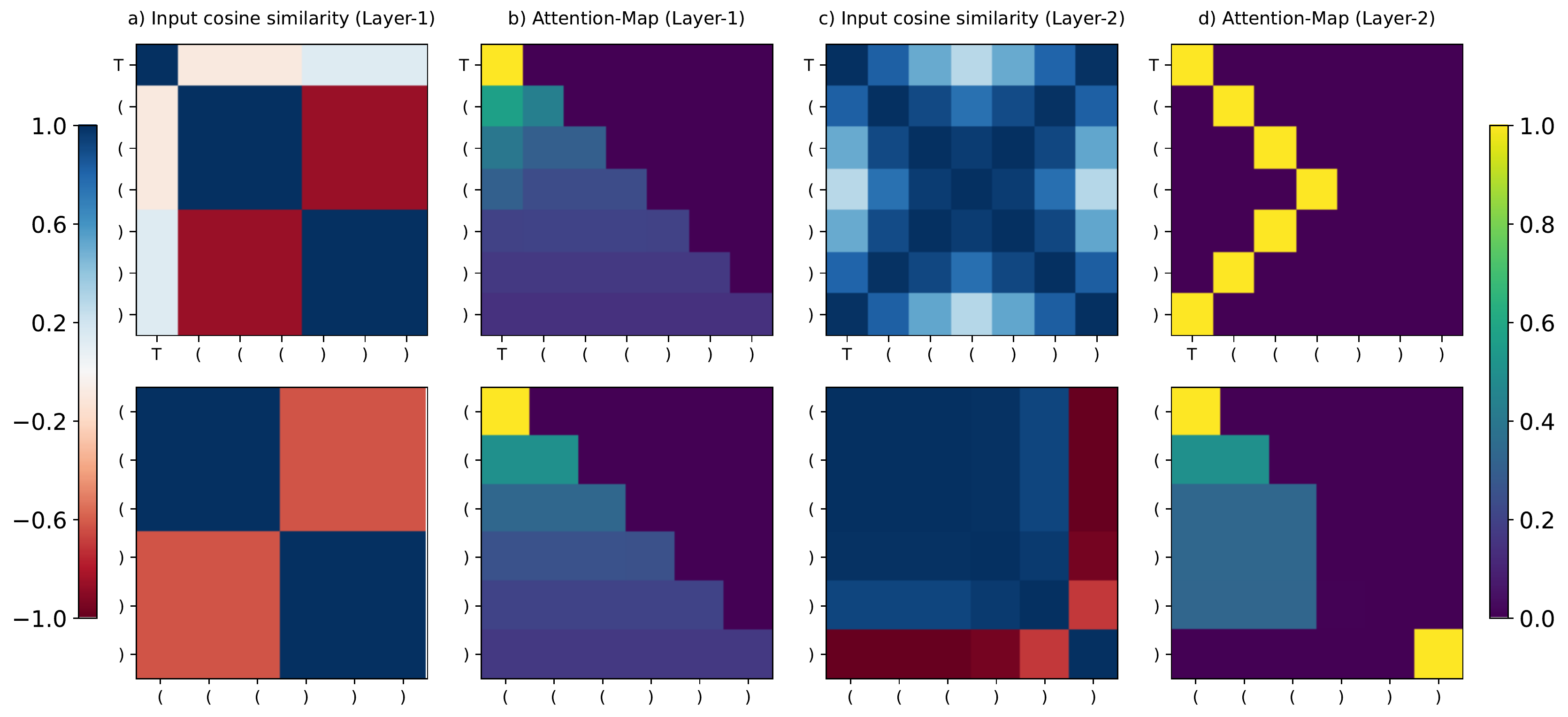}
  \captionof{figure}{Contrasting SA$^+$ and SA$^-$ on different parts of the network for the sequence ``((()))''. In a and c, we present the pair-wise cosine similarity of the symbols at layers 1 and 2. In b and d, we present the weights given by one of the attention heads.}
  \label{fig:vis_interpretability}
\end{strip}
\newpage


We take our SA$^+$ and SA$^-$ models trained on $\mathcal{D}_4$, and contrast the representations learned across the two layers of the network. For presentational purposes, we use a simple string, ``'((()))''.  Figure \ref{fig:vis_interpretability} shows the cosine similarity between pairs of symbols in the sequence and the attention-maps for the first (out of 4) heads at the two layers of the networks. The input sequence at the first layer is simply the looked-up embeddings for each symbol, which are identical for all opening parentheses, and similarly identical for all closing parentheses. The embeddings for the opening and closing parenthesis have negative cosine similarity: -0.77 for SA$^-$ and -0.95 for SA$^+$. Further, the starting symbol in SA$^+$ has a negative cosine similarity (-0.19) with the opening parenthesis and a positive cosine similarity (0.18) with the closing parenthesis. For both models, attention weights at the first layer are almost uniformly distributed across the preceding parentheses, opening or closing. This occurs because the input sequence to the first layer contains several identical representation for opening and closing parentheses. 

Beyond the first layer, the two networks behave radically differently. For SA$^-$, the the input representations to second layer have a cosine similarity close to or exactly 1.0, except for the last symbol. In contrast, the input representation for SA$^+$ is based on the head-dependency relationship. For instance, each opening parenthesis has the highest cosine similarity with its opening counterpart and the last closing parenthesis is matched with the starting symbol. Crucially, the starting symbol has enabled SA$^+$ to differentiate among the opening parentheses, which remain identical at layer-2 for the SA$^-$ model.
Both SA$^-$ and SA$^+$ maintain opposite representations (negative cosine similarity) for opening and closing parentheses, which helps them emulate push/pop operations. But only SA$^+$ is able to refine representations at the second layer, such that it can match the correct pair of opening and closing parentheses.


 
 
\end{document}